\documentclass{article}
\usepackage{ijcai24}
\usepackage{verbatim}
\usepackage{times}
\usepackage{soul}
\usepackage{url}
\usepackage[utf8]{inputenc}
\usepackage[small]{caption}
\usepackage{graphicx}
\usepackage{amsmath}
\usepackage{amsthm}
\usepackage{booktabs}
\usepackage{algorithm}
\usepackage{algorithmic}
\usepackage[switch]{lineno}
\usepackage{multirow}
\usepackage{amssymb}
\usepackage{graphicx}
\usepackage{amsmath}
\usepackage{tablefootnote}
\theoremstyle{definition}

\usepackage{dblfloatfix}
\usepackage{verbatim}
\usepackage{float}

\usepackage{enumitem}
\usepackage[pagebackref,breaklinks,colorlinks]{hyperref}
\usepackage[capitalize]{cleveref}
\crefname{section}{Sec.}{Secs.}
\Crefname{section}{Section}{Sections}
\Crefname{table}{Table}{Tables}
\crefname{table}{Tab.}{Tabs.}

\usepackage{subcaption}
\title{Using joint angles based on the international biomechanical standards for human action recognition and related tasks}

\author{
Kevin Schlegel $^1$\thanks{Work done while the author was a Research Fellow at University College London.}
\and
Lei Jiang$^{2}$\and
Hao Ni$^{2}$\thanks{Corresponding author. Email address: h.ni@ucl.ac.uk (Hao Ni)}\\
\affiliations
$^1$ Vicon Motion Systems, Oxford, UK\\
$^2$ University College London, London, UK\\
}

\begin{document}
\maketitle
\begin{abstract}
Keypoint data has received a considerable amount of attention in machine learning for tasks like action detection and recognition. However, human experts in movement such as doctors, physiotherapists, sports scientists and coaches use a notion of joint angles standardised by the International Society of Biomechanics to precisely and efficiently communicate static body poses and movements. In this paper, we introduce the basic biomechanical notions and show how they can be used to convert common keypoint data into joint angles that uniquely describe the given pose and have various desirable mathematical properties, such as independence of both the camera viewpoint and the person performing the action. We experimentally demonstrate that the joint angle representation of keypoint data is suitable for machine learning applications and can in some cases bring an immediate performance gain. The use of joint angles as a human meaningful representation of kinematic data is in particular promising for applications where interpretability and dialog with human experts is important, such as many sports and medical applications. To facilitate further research in this direction, we will release a python package to convert keypoint data into joint angles as outlined in this paper.\footnote{\label{foot:code}A GitHub repository will be made available later.}

\end{abstract} 
\section{Introduction}
\label{sec:intro}
Action recognition and more generally understanding human motion from data has received a lot of attention in machine learning due to its wide applications, including human-robot interaction, virtual reality and video surveillance~\cite{liu2019rgb}. Among the various data modalities one can consider for this task, skeleton data has received an increasing amount of attention recently~\cite{liao21,presti16,shi19,yan18,zhu2021dyadic,jiang2024gcn,liu2022symmetry} as this kind of data has become more readily available with the advent of powerful deep learning systems for pose estimation~\cite{cao2017realtime} and specialised hardware such as the Microsoft Kinect. Skeleton data provides many advantages over RGB video, such as a decoupling of the movement information from other information contained in the video, a considerable decrease in the size of the data and de-identification of the data, which can be crucial in many applications. However, while humans can easily recognise a movement from viewing the skeleton data, it is not very intuitive for us to describe a movement in this way. In particular in the medical and sports science communities, which have expert knowledge of biomechanics, it is standard to describe a pose or movement in terms of joint angles. Since the 1980s the International Society of Biomechanics (ISB) and other organisations have worked on creating standards for reporting kinematic data in order to facilitate communication and comparison of data and results between different groups \cite{wu95,wu02,wu05,grood83}. These standards are widely used in many contexts where a pose or movement needs to be communicated clearly and precisely, e.g.\ in medical journals studying muscle recruitment during exercise \cite{willcox13} or to describe the ideal positioning of a joint during surgery \cite{knight04}. In sports and exercise research the very first step after collecting motion capture data in a research study is often to translate this data into joint angles. However, these notions of joint angles have not yet been applied much in machine learning and data science, despite a number of further advantages compared to skeleton data, which include
\begin{enumerate}[itemsep=0.5pt]
    \item Joint angles are viewpoint and subject independent
    \item Further de-identification (removing e.g.\ height and gender-specific skeletal differences)
    \item Joint angles are human meaningful, in particular potentially benefit explainability of e.g.\ medical applications
    \item Individual movements are decoupled, the angles at each joint are independent of the angles elsewhere
    \item Data reduction (on the order of 1/3 of 3D keypoint data)
\end{enumerate}
This might be due to a lack of familiarity with the anatomical and biomechanical concepts and some challenges which need to be addressed to optimize the use of the standards defined by the ISB for computational purposes while preserving the human intuition. With this paper we aim to give an introduction into some of the fundamental biomechanical concepts and provide a starting point to investigate the use of joint angles in machine learning applications. We introduce a notion of joint angles based on local coordinate systems, which is consistent with the ISB standards and expert intuitions and has desirable mathematical properties, such as being a bijective mapping so that one can recover the skeleton data from its angle representation (assuming knowledge of the length of individual bones). This should be beneficial in a variety of applications, especially in the medical and sport science area, such as assessing posture and quality of movement for injury prevention and rehabilitation as well as technique coaching.

We will first give a general introduction into the relevant anatomical and biomechanical terms in \cref{sec:anatomy} and subsequently carefully describe the definition of local joint coordinate systems and joint angles we propose to use for computational applications in \cref{sec:jcs}. We will discuss challenges to be addressed or trade-offs to be made as and when they arise. Most notably, we only consider a subset of the joint coordinate systems one may find in the standards by the ISB. This is because many of the coordinate systems defined by the ISB are beyond the level of detail found in common sets of keyoints. The choice of coordinate systems we consider in this paper is exactly so that all the information contained in the keypoint data commonly used in machine learning research is represented. Alongside this paper, we release a python package to compute joint angles as described in this paper from keypoint data for a variety of different keypoint sets, in particular including all of the most common ones such as collected by a KinectV2 or OpenPose pose estimation. Moreover, the package allows to recover skeleton data from the joint angle representation.\footnotemark[\value{footnote}]

To show that joint angles can be used as an alternative representation of kinematic data we perform a range of experiments comparing the performance of various standard deep learning models on both joint angle data and keypoint data. For a direct comparison of the data representations we primarily compare the performance of standard time series classification methods such as LSTMs and transformers on both keypoints and angles without any adjustments to the spatio-temporal structure of kinematic data, but we also give a comparison of these standard models to specialised skeleton based human action recognition (SHAR) models from the literature. Further, we perform experiments on various datasets of varying data quality, ranging from high quality motion capture data, to data from pose estimation and the Microsoft Kinect. The results are presented in \cref{sec:experiments}.

The main contributions of this paper are:
\begin{enumerate}[itemsep=0.5pt]
    \item Introduce the notion of joint angles following biomechanical standards, which are meaningful to human experts and with important mathematical properties, in particular a one-to-one correspondence between a given body pose and a set of joint angles, continuity of angles with respect to movement at the joint almost everywhere and independence of viewpoint and subject.
    
    \item We perform numerical experiments to demonstrate that the joint angles contain the relevant information to make meaningful predictions, by comparing performance in the task of action recognition. These experiments support that incorporating our proposed angle representation can complement the effectiveness of keypoint representation, leading to a significantly improved performance in terms of accuracy and robustness against rotation.   

    \item Compared with other angle representations based on the standard dot product, our proposed joint angle representation contains more meaningful information, delivering significantly better prediction results.  
\end{enumerate}

\section{Anatomical terms of motion}
\label{sec:anatomy}
\begin{figure*}
  \centering
    \includegraphics[width=0.7\linewidth]{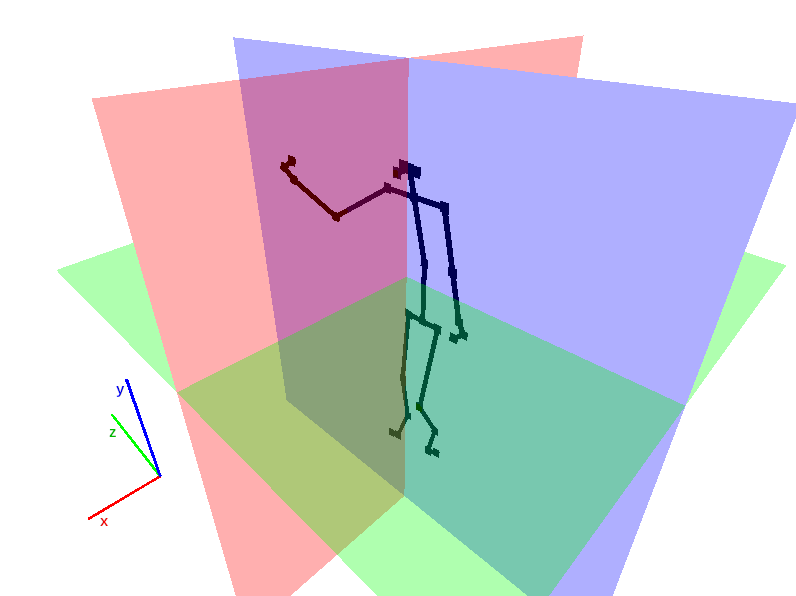}
   \caption{The three planes of motion: sagittal (xy, red), frontal (yz, blue) and transverse (xz, green).}
   \label{fig:movementplanes}
\end{figure*}

Most movements in the body can be decomposed into at most three distinct components, which are characterised by the plane in which the movement happens. Broadly speaking, for a coordinate system where the y axis points up relative to the anatomic position of the body, the x axis points forward and z points to the side away from the body, the three planes of movement are the \textit{sagittal} (xy), \textit{frontal} (yz) and \textit{transverse} (xz) planes (\cref{fig:movementplanes}). Thus the sagittal plane is comprised of the directions one would think of as forward \& backward and up \& down. The frontal plane is given by the left \& right and up \& down directions and the transverse by left \& right and forward \& backward.

Now, decomposing and describing movements using these three planes of motion, first, movements in the sagittal plane, moving the body or limb anteriorly or posteriorly, are referred to as \textit{flexion} and \textit{extension} with flexion directed forward and extension backward. These can also be movements affecting the angle between two adjacent bones (e.g.\ at the elbow) with flexion decreasing the angle and extension increasing the angle.

Second, movements in the frontal plane, moving a limb laterally towards or away from the body are referred to as \textit{abduction} and \textit{adduction}, with abduction being a movement away from the midline of the body and adduction towards it. For bodyparts that are on the midline of the body like the spine and neck any movement away from this midline to the side of the body if referred to as \textit{lateral flexion}.

Finally, movements in the transverse plane are generally called \textit{internal and external (axial) rotations}, with the direction again determined by whether the rotation is towards or away from the midline (unless the body part is on the midline).

While these definitions are mostly global, by placing an individual local coordinate system at each joint, most joint movements can be precisely described locally using these three types of movement. This way an individual joint position or movement can be specified independent of the state of any other joint within the body or the bodys global orientation in space. For this reason this way of reporting body pose or movement is standard in the medical and sports science communities.

\subsection{Limitations}
Before describing in detail how one can use the above terminology to precisely describe a pose or movement we need to discuss some of the limitations and simplifications in what we described. As stated above, the majority but not all movements of the human body can be characterised as described. The large scale movements which are primarily relevant for high-level tasks such as action recognition all do fall into these categories. The movements that we will not be discussing in this paper generally are missing because common skeleton datasets lack the keypoints for these movements to be visible. In particular, the single shoulder keypoint in skeleton datasets essentially corresponds to the glenohumeral joint (the ball-and-socket type joint between the humerus (upper arm bone) and the shoulder complex) so that we cannot see e.g.\ the acromioclavicular and sternoclavicular joints (either end of the clavicle) and cannot include protraction/retraction, elevation/depression and up-/downward rotation of the scapula. Similarly, the lack of keypoints along the spine means movement within the cervical, thoracic or lumbar spine cannot be detected. Creating new datasets which include the relevant extra keypoints would open up a wide range of new interesting applications, such as assessing posture and assessing quality of movement in strength training and rehabilitation exercises.\\

Aside from missing keypoints there is a difficulty in precisely describing the position of a joint which can express flexion and abduction. Flexion is an elevation parallel to the sagittal plane, i.e.\ a rotation around the local x axis and abduction is an elevation in the frontal plane, i.e.\ a rotation around the local z axis.
These rotations are not commutative and so performing flexion followed by abduction will in general give a different result than abduction followed by flexion (\cite{anglin00}, Section 8.1). In order to be able to use a vector of joint angles to uniquely describe a pose we propose to define the joint angles akin to spherical coordinates, so that every pair of flexion and abduction angles uniquely determines the position of the moving segment of the joint but human intuition is preserved, i.e.\ a practitioner familiar with these notions of joint angles, knowing the precise choices we made, could estimate the joint angles from viewing a skeleton and estimate a pose from reading a given set of angles. We will describe this in more detail below.

\section{Joint Coordinate Systems}
\label{sec:jcs}

\begin{figure*}[t!]
  \centering
  \begin{subfigure}{0.32\linewidth}
    \includegraphics[width=0.99\linewidth]{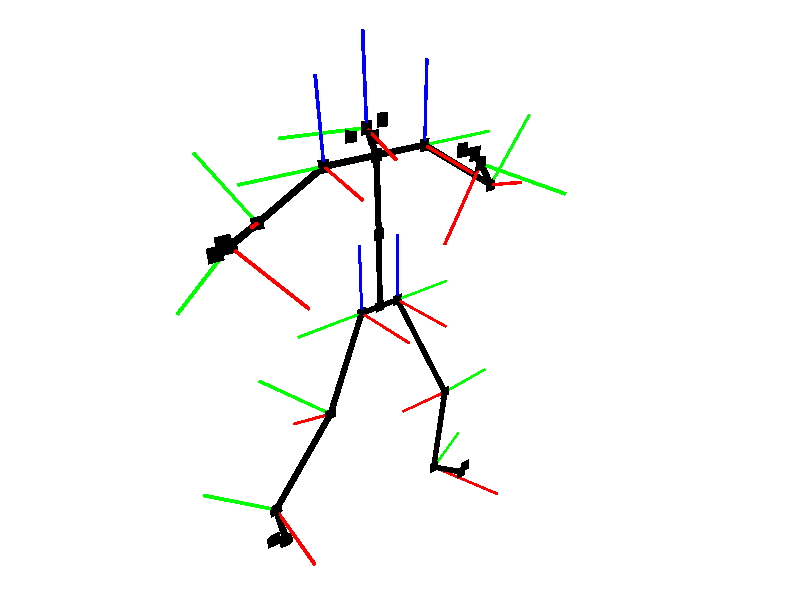}
    \caption{A local coordinate system is placed at each joint to describe the position of the moving segment of the joint independent of other joints and global viewpoint.}
    \label{fig:jcs}
  \end{subfigure}
  \hfill
  \begin{subfigure}{0.30\linewidth}
    \includegraphics[width=0.99\linewidth]{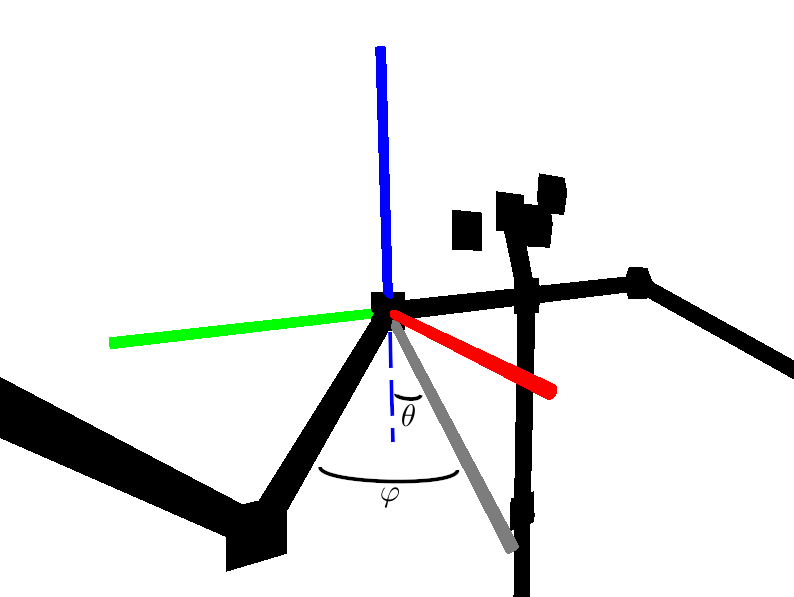}
    \caption{Shoulder angles: The xy projection of the arm is shown in grey. Flexion is measured as the angle between the projection and the negative y axis (blue) and abduction is measured as the angle between the arm bone and the projection.}
    \label{fig:shoulder_angles}
  \end{subfigure}
  \hfill
  \begin{subfigure}{0.32\linewidth}
    \includegraphics[width=0.99\linewidth]{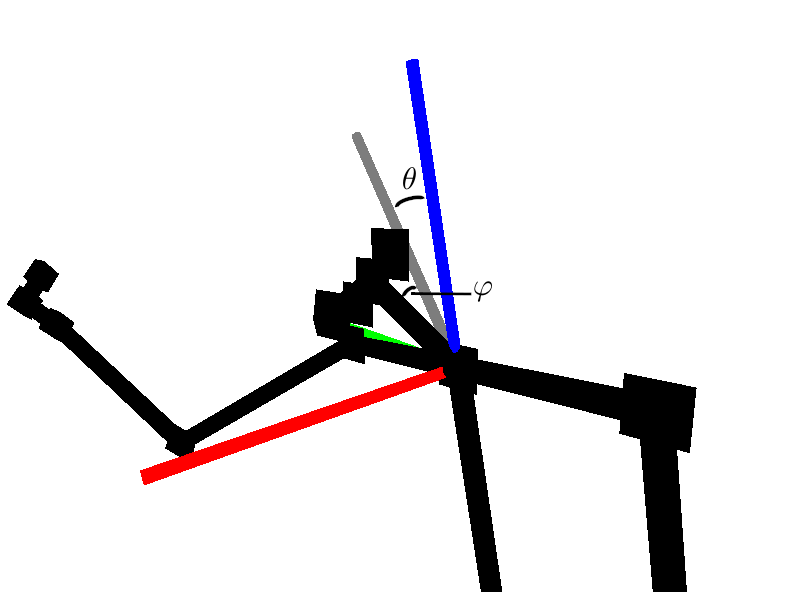}
    \caption{Neck angles: The yz projection of the head is shown in grey. Flexion is measured as the angle between the head and the projection and abduction is measured as the angle between the projection and the y axis (blue).}
    \label{fig:neck_angles}
  \end{subfigure}
  \caption{A visualisation of all JCS and the computation of shoulder \& neck angles.}
  \label{fig:joint_angles}
\end{figure*}



In order to precisely and efficiently describe a pose by a collection of joint angles we simplify and view each joint as a simple joint with only two articulation surfaces, i.e.\ a joint consists of one moving segment which moves with respect to the position of a fixed segment (e.g.\ at the hip the thigh bone moves with respect to the pelvis bone). We then place a local joint coordinate system (JCS) at each joint. It is clear that in order to determine the position of the moving bone in space the orientation of the JCS can only depend on the fixed segment and potentially movement upstream, and must be independent of the moving segment. As in the global view, the x axis generally points forwards, the y axis up and the the z axis to the side, away from the midline of the body. For joints which are located on the midline itself the ISB standards define the convention of the z axis pointing to the right hand side. \cref{fig:jcs} shows a skeleton with the local JCS at each joint.

We can then decompose most movements as a combination of the three major types described above. Flexion and abduction define where the moving segment is positioned in space with respect to the fixed segment and axial rotation is a rotation of the moving segment around its own axis, thus not affecting the position of that bone in space but the orientation of the JCS at the other end of that bone. Not all joints are able to articulate movement in all three planes. Which movements can be expressed depends on their structure but in every case the set of at most three angles defines the state of the joint exactly. By computing the joint angles with respect to the local JCS the joints position is described independent of both global factors (camera angle and the orientation of the body in space) as well as the pose at any other joint in the body.

\subsection{Specific definition of each joint coordinate system}
To define a local joint coordinate system (JCS) for each joint we first convert a given set of keypoints into a bone structure such as depicted in \cref{fig:jcs}. At each joint there is then exactly one bone which articulates the movement of the joint. It is clear that in order to be able to use the local JCS to determine the position of the moving bone in space the definition of the JCS has to be independent of this moving bone. Thus the JCS is defined using two different bones in the proximity of the joint which are not involved in the movement. The only exceptions to this rule are the elbows and knees where there are no three bones associated with the joint to define one moving and two static bones. At these joints, and only at these, we do use the moving bone to define the JCS. This is however not a problem and the JCS remains independent of the movement of the moving bones at these joints (the forearm and lower legs) because these bones can only move within the sagittal (xy) plane and we use the moving segment to define the z axis. For every JCS (including elbows and knees) one of its axes is always given by one of the static bones. A second axis is then set to be the cross product of this first axis with the second static bone (or in the case of elbows and knees, the cross product of the one static bone with the moving bone). The third axis always is simply the cross product of the first two axes. The order of axes, i.e. which of x,y and z axis are defined as either a static bone, the axis-bone or axis-axis cross product differs across different joints.

In the following we will describe in detail how the JCS for each joint is computed, i.e. which bones are used to define which axis in the way outlined above. For the hip joints we set the z axis to be the vector pointing from the opposite hip keypoint to the given hip keypoint. The x axis is computed as the cross product between this z vector and the bottom section of the spine, the connection from the pelvis to the closest keypoint along the spine. The shoulder JCS are defined similarly with the shoulder girdle vector pointing from one shoulder keypoint to the other and the upper section of the spine. We also define an upper and a lower proximal JCS, located at the centre of the shoulders and the pelvis respectively, which capture the state of the upper and lower section of the torso. Due to the lack of shoulder movement with current keypoint systems these two JCS are exactly the same as the right shoulder and hip JCS (following the right-handed convention by the ISB). For the elbows the y axis is given by the upper arm bone pointing from the elbow to the shoulder. The z axis is computed as the cross product between the upper and forearm bones. The knee JCS are computed similarly using the thigh and lower leg bones (Note that while the forearm and lower leg bones are the bones that articulate the movement at the elbow and knee they can be used to define the z axis as the cross product with the proximal bone of the limb because the knee and elbow can only move in the sagittal (xy) plane). For the JCS at the wrists the y axis is given by the forearm bone, the x axis is set to be the cross product of the forearm bone with the thumb (this is imprecise due to the range of motion of the thumb but the best approximation possible with common keypoint sets). For the ankles the y axis is given by the lower leg bone and the z axis by the cross product by the vector pointing from the base of the foot to the toes. Finally, the y axis of the neck JCS is set to the top section of the spine and the x axis as the cross product of the vector from one ear to the other, which appears to be the most stable set of keypoints to determine the orientation of the head.

\subsection{Joint angles}
Having defined a local JCS for each of the joints visible in common keypoint datasets we can now use these coordinate systems to precisely describe the position of the moving segment of each joint. The collection of all angles at all these joints then exactly encodes the pose of the given skeleton, so that with the additional knowledge of the length of each segment (which is personal information and not relevant for recognising a movement) the skeleton can be recovered from the set of joint angles.

As stated above, each joint has one bone which articulates the movements. In joints which can articulate axial rotation that rotation is around the moving segments length axis and thus only manifests itself downstream of the kinetic chain (since in keypoint data we cannot see the rotation of a bone around its own axis). Thus flexion and abduction (if articulated by a given joint) completely determine the position of the moving segment in space.
For the elbows and knees, which cannot move in the frontal plane, we can determine flexion simply as the angle between the moving bone (the forearm or lower leg) and the proximal bone (the upper arm or thigh) of the joint which also is the y axis of the joints JCS.
For the shoulders and hips, which can move in both sagittal and frontal plane, we define these two movements similar to a spherical coordinate system with the z axis as the zenith direction and with abduction similar to the polar angle $\varphi$ and flexion the azimuthal angle $\theta$. In contrast to typical spherical coordinates abduction is not the angle between the bone and the z axis, but the angle between the bone and its orthogonal projection on the sagittal (xy) plane. Flexion is measured as the angle of this orthogonal projection with the y axis. This is illustrated in \cref{fig:shoulder_angles}. With this definition pure flexion and pure abduction are precisely rotation around the z and x axis respectively. For a mix of flexion and abduction, where the order in which the limb was moved matters for the final position, we in a sense consider flexion to be the dominant movement. Abduction in this definition is then a rotation of the flexed bone around the y axis instead of a pure elevation in the frontal plane. However, in practise the results match intuition and uniquely identify the position of the moving bone in space. Moreover,  the angles are continuous with respect to the bone movement almost everywhere, except at exactly 90 degrees of abduction and 0 degrees flexion (when the moving bone aligns with the JCS z axis), which is important for computational use of these angles.
For all four joints we can determine their axial rotation by applying flexion and abduction (if applicable) to their JCS so that its y axis is aligned with the downstream JCS (e.g.\ align the shoulder JCS y axis with the elbow JCS y axis). Axial rotation can then directly be measured as the rotation of the two JCS against each other. This is in line with the ISB standards, where it is common to define two JCS for given joints which are associated with the ends of the bones forming the joint. The rotated shoulder JCS in the example would then correspond to the distal shoulder JCS and our elbow JCS corresponds to the proximal elbow JCS. This in particular means that axial rotation at the elbow and knee can only be determined when keypoints beyond the wrist and ankle are present, which determine the orientation of the wrist and ankle JCS.

Because of the structure of the joints, for the wrists and neck we reflect the movement of the joint best by using a spherical coordinate system with the x axis as zenith direction (different from the shoulder and hip JCS). Here we set abduction (or lateral flexion for the neck) as the polar angle $\varphi$ to be the angle between the bone and and its orthogonal projection on the frontal (yz) plane and flexion as the azimuthal angle $\theta$ to be again the angle between the bone and the y axis as shown in \cref{fig:neck_angles}. Axial rotation at the neck can be directly measured as the rotation of the neck JCS against the upper proximal JCS.

The relatively small range of motion and structure of the ankle joint means we can measure flexion very easily by directly measuring the angle between the leg and foot. Abduction on the other hand is only visible if we have a vector pointing across the foot such as the toe line to see its orientation. In that case we can measure abduction directly measuring the angle between this vector and the z axis of the ankle JCS.

Finally, the spine can articulate movement along most of its length, which is not well captured by common keypoint sets. In most cases there are no keypoints between the hips and shoulders so it is impossible to recognise any flexion. In this case we can determine lateral flexion and axial rotation simply by comparing the y and x axes respectively of the upper and lower proximal JCS. If we do have a keypoint along the spine between the shoulders and hips we can define all three movement angles, but the various ways the spine can move are still not well captured so to obtain intuitive results the definitions differ a bit from other joints. Disregarding translation in space axial rotation of the spine is the only movement to move the upper proximal JCS x axis within the xz plane of the lower proximal JCS, so we can determine axial rotation by considering the angle of the projection of the upper proximal x axis into this plane. Similarly in this view flexion is the only movement of the upper proximal x axis out of the lower proximal xz plane. Thus we can measure flexion as the angle between the axis and its orthogonal projection. Finally, knowing axial rotation and flexion,  lateral flexion can simply be determined by comparing the image of the lower proximal JCS z axis under both transformations to the z axis of the upper proximal JCS.

\section{Experiments}
\label{sec:experiments}

\begin{figure*}[t!]
  \centering
    \includegraphics[width=0.99\linewidth]{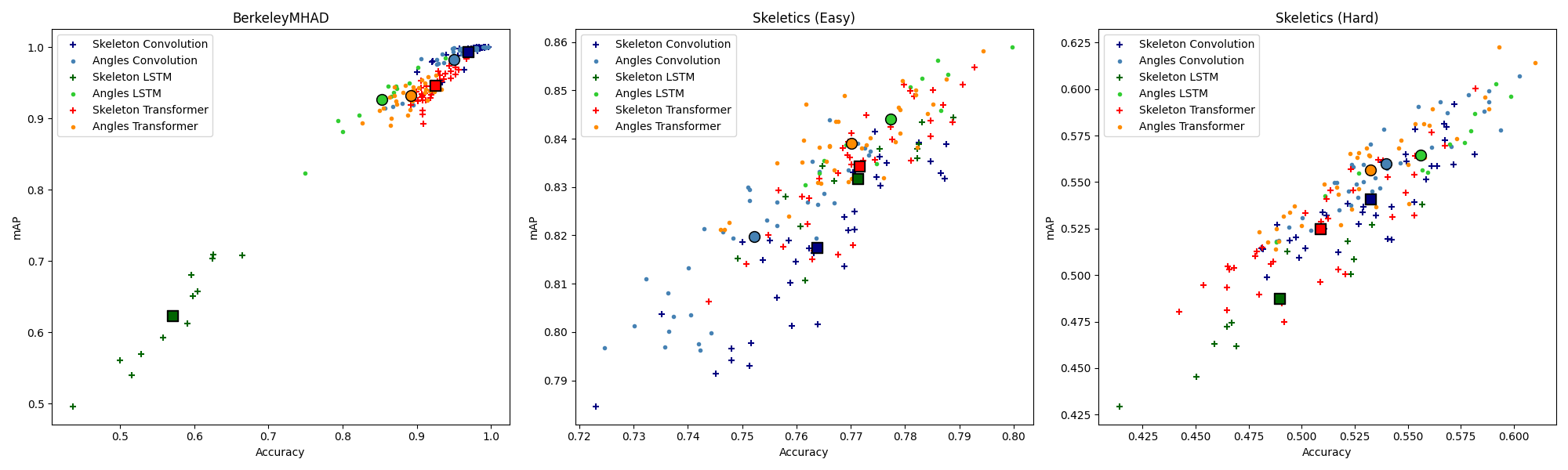}
  \caption{Performance comparison of joint angles and keypoints per dataset. Each small symbol represents one hyperparameter configuration with light dots for angles and dark crosses for keypoints, and similar colours for each model class. Large circles and squares represent the average performance of a given model class on joint angles and keypoints respectively.}
  \label{fig:res_data}
\end{figure*}

It is clear from the definitions in \cref{sec:jcs} that the joint angles are independent of the camera viewpoint (both in rotation and distance) as well as the subject (natural variations in length of individual body parts are removed, thus also further de-identifying the data). Moreover, the definitions have been chosen so that there is a bijection between keypoints and joint angles. Thus it is clear that joint angles carry the same information with respect to pose and movement as keypoint data does, while providing some additional beneficial properties.
In this section we evaluate whether common machine learning models are able to extract the relevant information to make meaningful predictions from this alternative data representation. We compare the performance of a range of deep learning models using either joint angles or keypoint data as their input for the standard action recognition task on several well known skeleton action recognition datasets. A lot of work in recent years went into the design of models that can effectively capture the spatio-temporal structure of keypoint data. In order to compare performance between angles and keypoints fairly we primarily use standard models for time series classification, 1D convolutions, LSTMs and transformers, without any attempt to capture the individual structures of angle or keypoint data. Each of these models receive either a vector of all joint angles or a flattened vector of keypoints as input. However, we do also give a comparison to some strong graph convolutional neural network (GCN) type models which make use of the keypoint structure.

The three dataset we perform experiments on are Berkeley MHAD~\cite{ofli13}, Skeletics152~\cite{gupta2020} and NTU RGB+D~\cite{shahroudy2016}.
BerkeleyMHAD \cite{ofli13} is a very small dataset, collected using a motion capture system, so the data is highly accurate. The motion capture was recorded at 480Hz, to make results more comparable between datasets of different data quality we subsample the data to a frame rate of 30Hz.
Skeletics152 \cite{gupta2020} is a large scale skeleton based human action dataset derived from the larger RGB video dataset Kinetics-700 using the VIBE pose estimation model \cite{kocabas2019}, thus representing realistic real-world data.
NTU RGB+D \cite{shahroudy2016} and NTU RGB+D 120~\cite{liu2019ntu} are datasets of daily actions recorded using the Microsoft Kinect. They are currently the most widely used skeleton based action recognition dataset, however the quality of the data is significantly lower than what can be obtained using modern pose estimation models (see e.g.\ NTU-X \cite{trivedi2021}, an attempt to improve the quality of the original NTU RGB+D data using pose estimation).
For all experiments the individual keypoint or angle sequences are normalised to a fixed length of 200 frames (close to the average per sequence frames of all datasets) using linear interpolation and all training is done using the Adam optimizer \cite{kingma15}.

\subsection{Comparison of angles and keypoints}
\label{sec:comparison}

\begin{figure*}
  \centering
    \includegraphics[width=0.99\linewidth]{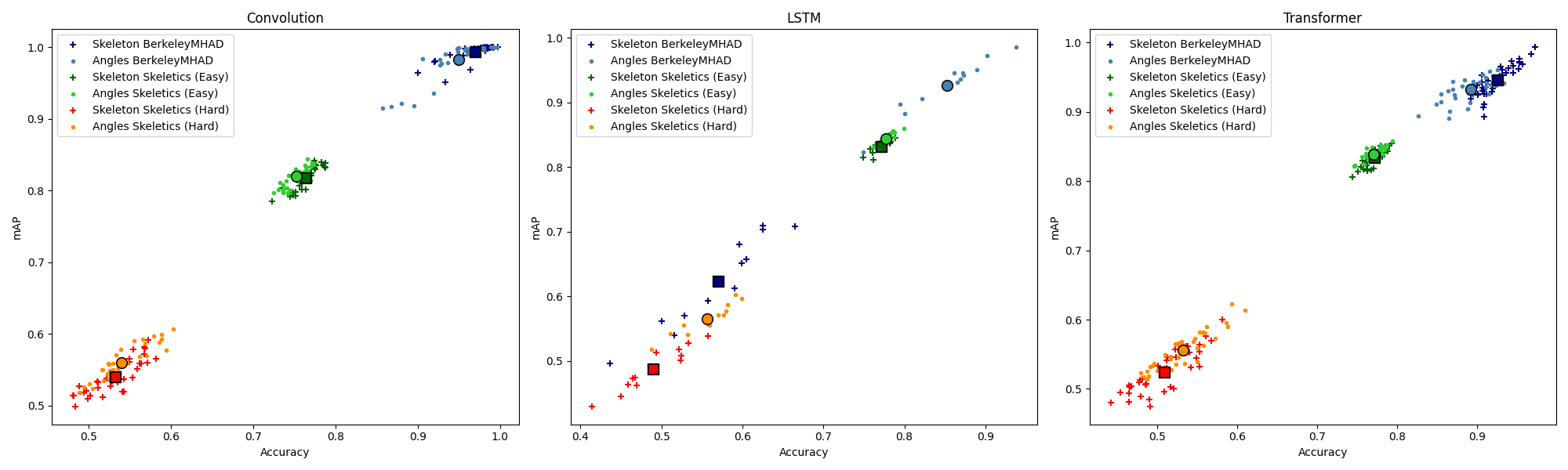}
    \caption{Performance comparison of joint angles and keypoints per model class. Each small symbol represents one hyperparameter configuration with light dots for angles and dark crosses for keypoints, and similar colours for each dataset. Large circles and squares represent the average performance on a given dataset on joint angles and keypoints respectively.}
  \label{fig:res_model}
\end{figure*}

We compare the performance of angles vs raw keypoint data using three standard time series classification models, 1D convolutions, LSTMs and transformers. For each model type the default PyTorch implementations are used. To ensure the results are not influenced by a particular choice of architecture hyperparamters we perform a small grid search for each (details of the hyperparamter grids can be found in \ref{appendix:hyperparamters}). We compare the performance using three different datasets. On the one hand we use Berkeley MHAD as it provides perfect data quality and on the other hand we use two subsets of comparable size (in the number of action classes) of the Skeletics 152 datasets for realistic real world data. The two subsets differ in the difficulty to distinguish their actions, with the first consisting of ten easy to distinguish classes such as squats, push-ups and pull-ups and the second one consisting of ten classes which are very similar in appearance, such as a range of throws and racket sports (a full list of the classes in each subset can be found in \ref{appendix:data}). We deliberately perform the experiments on small datasets and using small models (all the models used have at most $\sim$400k parameters), as one can expect the data representation to make a bigger difference in this case.

The results are shown per model and per dataset in \cref{fig:res_data,fig:res_model}. Broadly we see very similar performance on both angle and keypoint data, confirming that joint angles are indeed a viable representation of kinematic data to use as input for machine learning models e.g. to make use of its invariances under viewpoint and subject or to make in- and outputs more human meaningful.


On a more fine scale in \cref{fig:res_data} we can see a that while on the easy to distinguish, high quality BerkeleyMHAD data there is generally very little difference in performance between keypoints and angles (except for LSTMs), there is a gradual performance shift in favour of the joint angles as the dataset becomes more difficult. On the Skeletics subset with very distinct actions performance is still almost on par between the two representations, however on the Skeletics subset which contains many action classes with very similar poses we see a notable performance gain from using joint angles. This performance gain manifests in two ways, first each model class performs on average better using joint angles than using keypoints and second, the instance of each model class performing best on joint angles outperforms all other models of any class on keypoints. This suggests that on keypoint data the models learn to distinguish broadly different body poses but struggle to capture the more subtle differences that distinguish e.g. different types of throws or racket sports. These more subtle variations of movement are easier to capture using the notion of joint angles.


We further can see in \cref{fig:res_model} that while convolutions perform equally on both kinds of data, transformers perform better on skeletons on the high quality Berkley data, but better on angles on more difficult data, in particular on difficult to distinguish classes. Finally, LSTMs consistently perform better on angle data than on keypoint data. This points to the fact that the joint angles represent the spatio-temporal structure of kinematic data differently from keypoints and the model architecture can have a profound influence on the kind of structures in the data that can be captured well. A large amount of work has been done in recent years on how one can best capture the structures within keypoint data. It is likely that careful architecture design to effectively capture the structure of joint angle data will be similarly valuable to obtain the maximal benefit of the joint angle representation.

\subsection{Comparison with structured keypoint models}

\begin{table}[t]
  \centering
  \resizebox{\columnwidth}{!}{%
  \begin{tabular}{@{}lcccc@{}}
    \toprule
     & \multicolumn{2}{c}{Skeletics (Easy)} & \multicolumn{2}{c}{Skeletics (Hard)}\\
    Model & Acc. & mAP & Acc. & mAP \\
    \midrule
    Conv & 0.7876 & 0.8388 & 0.6028 & 0.6070 \\
    LSTM & 0.7998 & 0.8589 & 0.5986 & 0.5962 \\
    Transformer & 0.7944 & 0.8582 & \textbf{0.6102} & \textbf{0.6140}\\
    \midrule
    ST-GCN \cite{yan18} & 0.8075 & 0.8788 & 0.5725 & 0.6084\\
    AGCN \cite{shi19} & \textbf{0.8094} & \textbf{0.8798} & 0.5586 & 0.5865\\
    \bottomrule
  \end{tabular}}
  \caption{Comparison of standard models on joint angles to SHAR GCN-type models on keypoints.}
  \label{tab:gcn}
\end{table}

We also compare the performance of the standard models from the previous section performing best on the joint angle data with two strong GCN-type networks which have been designed to capture the spatio-temporal structure of keypoint data, ST-GCN \cite{yan18} and AGCN \cite{shi19}. The results are summarised in \cref{tab:gcn}. We see a similar pattern as above emerge, in that on the data with very distinct poses the GCN models on keypoint data clearly outperform the non-specialised models on joint angles. On the data where the differences between actions are more subtle however, almost every standard model on joint angles is able to outperform the GCN models on keypoints. It is worth pointing out again that these experiments were done on small datasets and we know that GCN type models in general need large amounts of data to exhibit their full potential. However, the small size of the datasets helps to demonstrate the effectiveness of the data representation. Moreover it is very relevant to many real world applications where obtaining large datasets can be very costly, as is in particular often the case in the medical field. To account for the small dataset sizes for these experiments we considered truncations of the two GCN models, taking only the first $n$ layers of the models presented in the respective papers. We tested various truncation levels and used the best performing one for the results in this section (3 layers for AGCN and 3 and 5 layers for ST-GCN on the easy and hard subset respectively).

\subsection{Dataset size}
\begin{figure}[t!]
  \centering
    \includegraphics[width=0.9\linewidth]{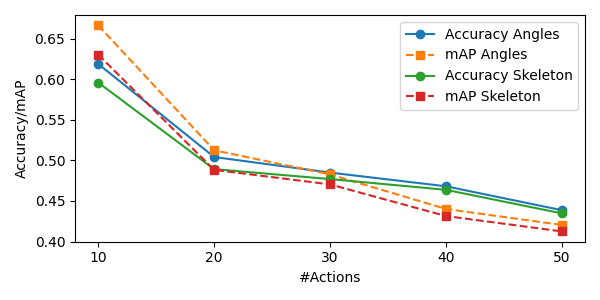}
  \caption{Performance comparison of transformers by dataset size.}
  \label{fig:res_size}
\end{figure}

With our experiments up to this point based on the intuition that differences in the data representation become more apparent in smaller datasets, here we evaluate the change in performance of the standard time series classification models from \cref{sec:comparison} as the dataset size increases. For this we compare the average performance of the same models as before on the first $n=10,20,30,40,50$ actions of the skeletics dataset. The results for transformers are shown in \cref{fig:res_size} (we omit the plots for other model classes as they show the same general behaviour). We find that as expected on larger datasets the performance gap decreases as the model has a greater chance of learning some of the invariances the joint angle representation provides. However, the performance of joint angles generally remains slightly superior to using raw keypoint data.

\subsection{Comparisons on large-scale NTU dataset}
\label{NTU}
To further demonstrate the superiority of our proposed angle representation based on the international biomechanical standards, we also conduct extensive experiments on the large-scale NTU RGB+D dataset that comprises 60 action classes with 56,880 samples and NTU-RGB+D 120 dataset that has 120 action categories with 114,480 examples. 

\subsubsection{Incorporating joint angles with keypoints}
\begin{figure}[h]
\centering
\setlength{\tabcolsep}{1pt}
\begin{tabular}{cc}

\includegraphics[width=0.5\linewidth]{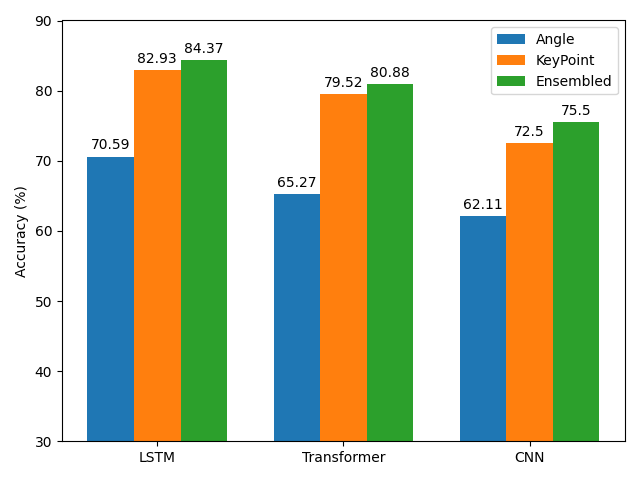} &
\includegraphics[width=0.5\linewidth]{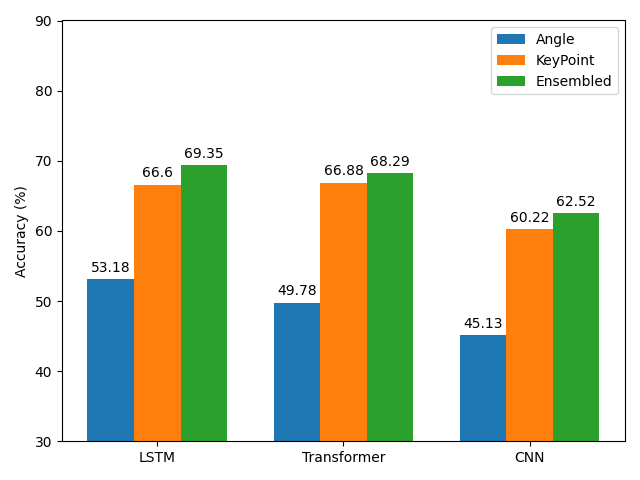}
\\
 \small{(a) NTU RGB+D}  &  \small{(b) NTU RGB+D 120}  \\
\end{tabular}
\caption{The accuracy of keypoint, angles and their ensemble using different models on (a) NTU RGB+D dataset and (b) NTU RGB+D 120 dataset.}
\label{ensemble_result}
\end{figure}

As illustrated in \cref{ensemble_result}, although the model employing our joint angles underperforms compared to the model using keypoint on NTU RGB+D or NTU RGB+D 120 dataset due to the poor dataset quality, detailed in \ref{appendix:quality}, ensembled result summing the results from the angle-based model and keypoint-based model always achieves the best performance on both datasets. It indicates that incorporating angle information can significantly enhance model efficacy. Extra results can be seen in \ref{appendix:ensemble}.


\subsubsection{Robustness against rotation}
\begin{table}[t]
  \centering
  \resizebox{\columnwidth}{!}{
  \begin{tabular}{l|cc|cc|cc}
  \hline
   & \multicolumn{2}{c|}{Keypoint}  & \multicolumn{2}{c|}{Joint angles}& \multicolumn{2}{c}{Ensembled} \\
   & Acc. & mAP & Acc. & mAP& Acc. &mAP \\
   \hline
   Before rotation & 0.8293 & 0.8862 &0.7059& 0.7613&0.8437&0.8942\\
    rotate $\pm$ 0.1 rad& 0.8209& 0.8803&0.7059& 0.7613&0.8383&0.8913\\
    rotate  $\pm$ 0.2 rad& 0.8013& 0.8607&0.7059& 0.7613 &0.8286&0.8823 \\
    rotate $\pm$ 0.3 rad& 0.7628& 0.8222&0.7059& 0.7613& 0.8075&0.8668 \\
    rotate $\pm$ 0.8 rad & 0.4730 &0.5070 &0.7059&0.7613& 0.7047&0.7659 \\
  \hline
  \end{tabular}
  }
  \caption{Robustness test of the rotation for models based on the keypoint, joint angles on NTU RGB+D dataset. }
  \label{tab:NTU}
\end{table}
One of the main advantages using angle representation is its rotation-invariant property, which means the model will keep high performance even though the skeleton joints are rotated due to noises or rotated camera views. In order to test the rotation robustness of the model based on the angle and keypoint data respectively, following~\cite{huang2024fusing}, we randomly rotate original keypoint within a certain radian to get the new keypoint data. In \cref{tab:NTU}, we show the LSTMs model performance before and after the rotation. Since angle representation is rotation-invariant, the angle derived from the new keypoint data after rotation remain same with the original keypoint. As the result, the model using angle as input maintain the same performance while the model utilising keypoint will perform significantly worse, with the accuracy dropping from 82.93\% to 47.3\% and mAP dropping from 88.62\% to 50.70\% when rotation angle reaches $\pm$0.8 radian. Again, Incorporating the results from joint angles can help ensembled system (summing results from the angle and joint based model) maintain high performance and robustness against rotation noises.        

\subsubsection{Comparison with other angle representations}
\begin{figure}[h]
\centering
\setlength{\tabcolsep}{1pt}
\begin{tabular}{cc}

\includegraphics[width=0.5\linewidth]{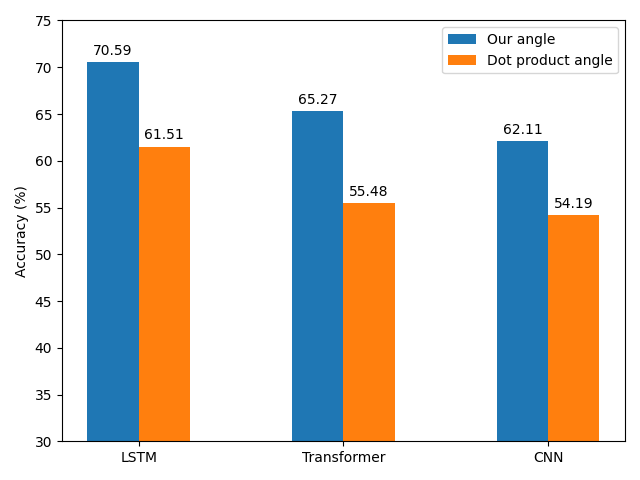} &
\includegraphics[width=0.5\linewidth]{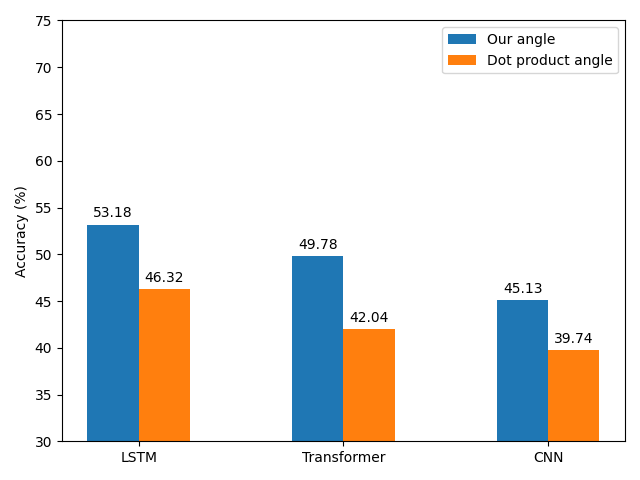}
\\
 \small{(a) NTU RGB+D}  &  \small{(b) NTU RGB+D 120}  \\
\end{tabular}
\caption{The accuracy comparison between our proposed angle representation and the angle obtained by the dot product on (a) NTU RGB+D dataset and (b) NTU RGB+D 120 dataset.}
\label{fig:angles}
\end{figure}
Although angle representation is also utilised in other approaches~\cite{huang2024fusing,qin2022fusing}, the computation of angles in these works simply follows the dot product of two vectors in $\mathbb{R}^3$, as shown below:
\begin{equation}
    cos \theta = \frac{u\cdot v}{|u||v|}
\end{equation}
In contrast, following international biomechanical standards, our proposed angle representation decomposes the body pose and movement onto three planes of motion and consider flexion, abduction and axial movements, capturing significantly more meaningful information. It is evident from \cref{fig:angles} that our angle representation delivers much better results compared to those obtained from the standard dot product method on both datasets.


\section{Conclusion}
\label{sec:conclusion}
The aim of this paper is to introduce the notion of joint angles based on international biomechanical standards to the machine learning community and demonstrate that they are a useful alternative representation of landmark data. The joint angles as defined in this paper have a one-to-one correspondence to keypoint data and are viewpoint and subject independent. We have demonstrated that they carry the information needed to recognise actions with at least similar, often better, performance to keypoint data without careful architecture design. However, the model architecture can have a significant effect on performance and designing model architectures adapted to the structure of joint angles is one of many important future research questions. It appears that joint angles capture the essence of a movement better, leading to consistently better performance when different actions have very similar overall poses.

Questions of explainability should also play a large role in the design of these architectures, as joint angles seem particular promising for applications in areas where we may want to understand the predictions of the model such as medical diagnoses, guiding rehabilitation work or coaching strength and conditioning training. For all these applications a human expert would communicate poses and movements in terms of joint angles, so would be able to interact with the software in a most natural way. In order to facilitate further research in this direction we release a python package to compute joint angles from many popular keypoint datasets.

\textbf{Acknowledgements.} LJ and HN are supported by the EPSRC [grant number
EP/S026347/1]. HN is also supported by The Alan Turing Institute under the EPSRC grant EP/N510129/1.
\bibliographystyle{named}
\bibliography{ijcai24}

\clearpage
\setcounter{page}{1}

\appendix
\section{Experimental details}
\subsection{Dataset choices}
\label{appendix:data}

We perform experiments on two subsets of the Skeletics 152 dataset. The first one of which consists of the following ten easy to distinguish action classes:\\

Squat, deadlift, snatch, clean \& jerk, lunge, push-up, mountain climber, front raise, pull up, battle rope\\

The second subset in turn consists of ten actions with very similar poses and movement patterns:\\

Playing tennis, playing ping pong, playing badminton, swinging baseball bat, catching/throwing softball, catching/throwing baseball, catching/throwing Frisbee, javelin throw, pull ups, climbing a rope\\

For our comparison with the NTU RGB+D dataset in~\ref{appendix:quality}, we used a similar subset of the original dataset, consisting of the following ten actions:\\

Pick up, throw, put on jacket, wave hand, kick something, jump up, point, sit down, stand up, salute

\subsection{Hyperparameter choices}
\label{appendix:hyperparamters}

We compare the performance of joint angles and keypoints using standard time series classification models, a convolutional model, an LSTM and a transformer. We use dimension = 128, layers = 3 and dropout = 0 for all experiments in \cref{NTU}. In other parts, for each model class we perform experiments on a small grid of hyperparameters, to make sure the results don't depend on a particular choice of hyperparameters. The hyperparameter grids for each class of model are as follows:\\

\textbf{Convolutions:}
\begin{itemize}
    \item Dimension of hidden state: $\{64,128\}$
    \item Kernel size: $\{3,5,7\}$
    \item Stride: $\{1,2\}$
    \item Layers: $\{3,4\}$
\end{itemize}
All convolutional models apply a dropout of 0.25 between any 2 convolutional layers.\\

\textbf{LSTM:}
\begin{itemize}
    \item Dimension of hidden state: $\{64,128\}$
    \item Transformer layers: $\{2,3\}$
    \item Dropout: $\{0,0.25,0.5\}$
\end{itemize}
\textbf{Transformer:}
\begin{itemize}
    \item Dimension of hidden state: $\{32,64\}$
    \item Number of attention heads: $\{2,4,8\}$
    \item Transformer layers: $\{2,3\}$
    \item Dropout: $\{0,0.1,0.25\}$
\end{itemize}

\section{Data quality}
\label{appendix:quality}
\begin{figure}[h]
  \centering
    \includegraphics[width=0.75\linewidth]{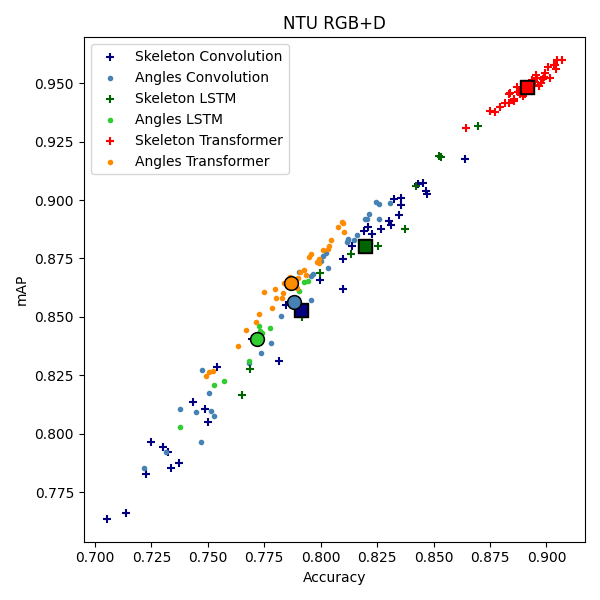}
   \caption{Performance comparison of joint angles and keypoints on a subset of NTU RGB+D.}
   \label{fig:ntu}
\end{figure}

Whenever we perform a transformation of raw data there is a question of stability. The Microsoft Kinect was one of the dominant sources of keypoint data for some time, due to its availability and low cost compared to motion capture and superior performance compared to early pose estimation models. However, examining of the data reveals that the Kinect data is generally very noisy and of significantly lower quality than modern pose estimation systems, as is for example demonstrated by the NTU-X project \cite{trivedi2021} aiming to improve the skeleton data of the NTU RGB+D dataset by using pose estimation. In order to evaulate the stability of the joint angle representation we performed the same experiments as presented in \cref{sec:comparison} on a subset of the NTU RGB+D dataset (the exact choice of actions can be found in \ref{appendix:data}), summarised in \cref{fig:ntu}. It is clear that on the low quality NTU data the performance of joint angles is consistently inferior to the use of keypoint data directly. This is likely another reason joint angles have not yet been adopted by the machine learning community, as datasets of sufficient quality have only recently started to emerge.

\section{Ensemble result}
\label{appendix:ensemble}
In this section, we demonstrate more results when combining the results from the keypoint-based model and joint angle-based model. Combining results from different models using distinct input representations has become a popular method to enhance the overall performance of a system in skeleton-based action recognition tasks since it was proposed by \cite{shi19}. In our work, we train two models separately using keypoints and our proposed joint angles. During evaluation, we sum their prediction probabilities for each class and select the final prediction with the highest combined probability. As we can see from \cref{tab:ensemble_NTU}, irrespective of the dataset or model architecture employed, the ensemble results by combining the outputs from keypoint-based model and joint angle-based model consistently have the best performance in terms of accuracy and mAP, suggesting that integrating angle representation can complement the effectiveness of keypoint representation.

\begin{table}[h]
  \centering
  \resizebox{\columnwidth}{!}{
  \begin{tabular}{c|c|cc|cc|cc}
  \hline
   \multirow{2}{*}{Dataset}  & \multirow{2}{*}{Model} & \multicolumn{2}{c|}{Keypoint}  & \multicolumn{2}{c|}{Joint angles}& \multicolumn{2}{c}{Ensemble} \\
   & & Acc. & mAP & Acc. & mAP& Acc. &mAP \\
   \hline
  \multirow{3}{*}{NTU RGB+D} &Conv &0.7250 & 0.7775 & 0.6211& 0.6653 &0.7550 & 0.8108\\
  &LSTM & 0.8293 & 0.8862 &0.7059 &0.7670  & 0.8437 & 0.8942\\
  &Transformer& 0.7952 & 0.8564 & 0.6527 &0.7139 & 0.8088 &0.8659  \\
  \hline

  \multirow{3}{*}{NTU RGB+D 120} &Conv & 0.6022 & 0.6227 & 0.4513 & 0.4616 & 0.6252 & 0.6613 \\
  &LSTM &  0.6660 &  0.7089 & 0.5318& 0.5621 & 0.6935 & 0.7343 \\
  &Transformer& 0.6688 & 0.7202  & 0.4978& 0.5235 &0.6829 & 0.7265\\
  \hline
  
  \end{tabular}
  }
  \caption{Comparisons for models based on the keypoint, joint angles and their combination on both NTU RGB+D dataset and NTU RGB+D 120 dataset. }
  \label{tab:ensemble_NTU}
\end{table}

\end{document}